\documentclass[10pt,twocolumn,letterpaper]{article}

\usepackage{wacv}
\usepackage{times}
\usepackage{epsfig}
\usepackage{graphicx}
\usepackage{amsmath}
\usepackage{amssymb}
\usepackage{booktabs}

\def\wacvPaperID{2175} 

\wacvfinalcopy 

\usepackage{times}
\usepackage{epsfig}
\usepackage{graphicx}
\usepackage{amsmath}
\usepackage{amssymb}
\usepackage{booktabs}
\usepackage{mathtools}
\usepackage{adjustbox}

\usepackage{verbatim}
\usepackage{multirow}
\usepackage{multicol}
\usepackage{makecell}
\usepackage{tabularx}
\usepackage{amsfonts}
\usepackage{floatrow}
\usepackage[normalem]{ulem}
\useunder{\uline}{\ul}{}
\usepackage{caption}
\usepackage{wrapfig}

\usepackage{algorithm} 
\usepackage{algpseudocode} 
\usepackage{tabto}

\usepackage{xcolor}
\usepackage{listings}
\usepackage[british,american]{babel}
\usepackage{enumitem}

\usepackage{textpos}  

\definecolor{mygray}{gray}{0.6}

\newfloatcommand{capbtabbox}{table}[][\FBwidth]
\DeclareCaptionFont{Small}{\fontsize{7}{7}\selectfont}
\DeclareCaptionFont{Ssmall}{\fontsize{6}{6}\selectfont}
\DeclareCaptionFont{Sssmall}{\fontsize{5}{5}\selectfont}

\newcommand{\app}{\raise.17ex\hbox{$\scriptstyle\sim$}}

\usepackage{tabulary}
\newcolumntype{x}[1]{>{\centering\arraybackslash}p{#1pt}}
\DeclarePairedDelimiter{\norm}{\lVert}{\rVert} 
\newcommand\inner[1]{\langle#1\rangle}
\ifwacvfinal
\usepackage[breaklinks=true,bookmarks=false]{hyperref}
\else
\usepackage[pagebackref=true,breaklinks=true,colorlinks,bookmarks=false]{hyperref}
\fi

\usepackage[capitalize]{cleveref}
\crefname{section}{Sec.}{Secs.}
\Crefname{section}{Section}{Sections}
\Crefname{table}{Table}{Tables}
\crefname{table}{Tab.}{Tabs.}
\Crefname{figure}{Figure}{Figure}
\crefname{figure}{Fig.}{Fig.}
\Crefname{equation}{Equation}{Equation}
\crefname{equation}{Eq.}{Eq.}

\begin{document}

\title{Self-Distilled Self-supervised Representation Learning}

\author{Jiho Jang$^1$,
Seonhoon Kim$^{2*}$,
Kiyoon Yoo$^{1*}$,
Chaerin Kong$^{1}$,
Jangho Kim$^{3}$,
Nojun Kwak$^{1}$\\\\
$^1$Seoul National University, $^2$Coupang, $^3$Kookmin University\\
{\tt\small\{geographic,961230,rin4616,nojunk\}@snu.ac.kr, sekim625@coupang.com, jangho.kim@kookmin.ac.kr}
}
%
%

\maketitle
\thispagestyle{empty}

\def\thefootnote{*}\footnotetext{Equal Contribution}\def\thefootnote{\arabic{footnote}}

\begin{abstract}
State-of-the-art frameworks in self-supervised learning have recently shown that fully utilizing transformer-based models can lead to performance boost compared to conventional CNN models. Striving to maximize the mutual information of two views of an image, existing works apply a contrastive loss to the final representations. Motivated by self-distillation in the supervised regime, we further exploit this by allowing the \textit{intermediate representations} to learn from the final layer via the contrastive loss. Through self-distillation, the intermediate layers are better suited for instance discrimination, making the performance of an early-exited sub-network not much degraded from that of the full network. This renders the pretext task easier also for the final layer, leading to better representations. Our method, Self-Distilled Self-Supervised Learning (SDSSL), outperforms competitive baselines (SimCLR, BYOL and MoCo v3) using ViT on various tasks and datasets. In the linear evaluation and k-NN protocol, SDSSL not only leads to superior performance in the final layers, but also in most of the lower layers. Furthermore, qualitative and quantitative analyses show how representations are formed more effectively along the transformer layers. Code is available at https://github.com/hagiss/SDSSL.
\end{abstract}
\vspace{-4mm}
\section{Introduction}

GPT \cite{radford2018improving} and BERT \cite{devlin2018bert} are two representative works in self-supervised learning (SSL) 
that use transformers \cite{vaswani2017attention} for natural language processing (NLP) tasks. 
Motivated by these successes, various efforts on self-supervised representation learning \cite{oord2018representation,hjelm2018learning,bachman2019learning, kaku2021intermediate, dwibedi2021little}  have been made in the vision domain as well, many of which follow the recent paradigm of instance discrimination that matches the representations of different views of the same image generated by separate augmentations \cite{chen2020simple,he2020momentum,grill2020bootstrap,caron2020unsupervised, chen2021exploring}. 
Recent self-supervised frameworks have focused on using transformer-based models such as ViT \cite{dosovitskiy2020image}, which has demonstrated superior performance over the conventional ResNet \cite{he2016deep} architectures.  
MoCo v3 \cite{chen2021empirical} and DINO \cite{caron2021emerging} achieved state-of-the-art performances using ViT in self-supervised learning. 
MoCo v3 investigated the learning instability of ViT and tackled this to enhance performance, while DINO exploited the characteristic{s} of ViT and proposed a unique MLP head to improve representation learning. 

\begin{figure}[t]
    {\caption{\footnotesize\textbf{Representations in hypersphere.} An illustration of representations of the student's low layer and the teacher's output on a hypersphere. Intermediate self-distillation loss explicitly shifts the representations of the low layer to the output representations.\vspace{-3mm}}\label{fig:repr}}
    {\includegraphics[width=0.95\linewidth]{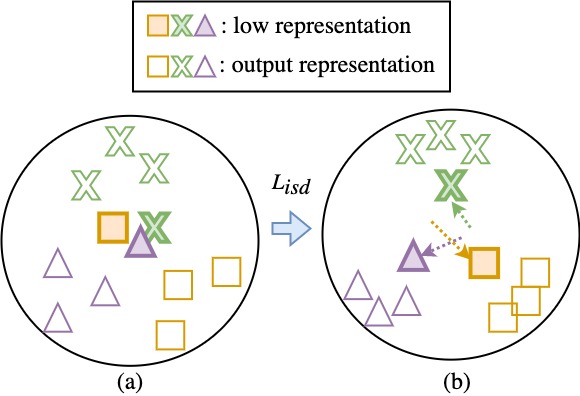}}
    \vspace{-2mm}
\end{figure}

Meanwhile, in the supervised regime, knowledge distillation via self-distillation \cite{zhang2019your}, which encourages the low layer outputs to follow the outputs of the final or higher layers, have shown to be effective with attempts to explain its performance boost by the mechanisms of ensemble \cite{mobahi2020self} and regularization \cite{allen2020towards}.  
Despite its effectiveness, self-distillation (SD) has not been utilized in the self-supervised framework with works focusing only on distillation between the \textit{final outputs} of a student and a teacher network (composed of exponential moving averages of students).\footnote{DINO uses the term ``self-distillation” to refer to distillation between a student network and a teacher network.} Motivated by this, we propose Self-Distilled Self-Supervised Learning (SDSSL), a natural application of SD to self-supervised learning.
When applied to SSL methods such as SimCLR and MoCo, self-distillation has an intuitive explanation: aligning representations of the lower layers to the final one can enhance linear separability of the lower layer representations as illustrated in Fig. \ref{fig:repr}. This in turn renders the instance discrimination task easier for the subsequent layers.

Consequently, by solving the instance discrimination pretext task better than its counterpart, we empirically demonstrate SDSSL increases the performance for multiple downstream tasks. In addition, we quantitatively and qualitatively show that SDSSL generates better intermediate representations. 

Because our method operates in an orthogonal manner to other SSL methods, we can simply apply our method on top of existing works.  
In this work, we apply our method to three representative SSL frameworks, namely SimCLR \cite{chen2020simple}, BYOL \cite{grill2020bootstrap}, and MoCo v3 \cite{chen2021empirical}, using ViT \cite{dosovitskiy2020image} as the backbone and show that our method improves upon the already competitive baselines. 
We demonstrate the effectiveness of SDSSL on ImageNet via k-nearest neighbor (k-NN) and linear evaluation. The superiority of SDSSL is also shown on various practical tasks such as copy detection, video segmentation and image retrieval.
We further investigate the representations learned by SDSSL using recently proposed metrics \cite{wang2020understanding} and discover that SDSSL makes a more linearly separable representation space compared to the baselines.
Finally, similar to \cite{phuong2019distillation,zhang2019your}, by encouraging the intermediate layers to explicitly learn the pretext task, we show that even the intermediate features can be successfully used for the downstream tasks, outperforming the baseline counterparts. 

Overall, we propose a self-distillation method that lets the intermediate layers \textit{explicitly} learn to discriminate instances. We show that our method improves upon the conventional SSL frameworks such as SimCLR, BYOL, and MoCo v3 on various benchmarks.
Through thorough ablation studies, we demonstrate that naively applying our method leads to performance degradation and show how our approach overcomes these potential pitfalls. 

\section{Related Work}
\noindent \textbf{Self supervised learning} Despite their impressive progress across multiple domains, deep neural networks (DNN) are extremely data-hungry, requiring a large scale dataset for training. As larger models demand even larger datasets, the annotation cost easily becomes unaffordable. For this reason, many works have explored the field of self-supervised learning, which is a family of unsupervised learning frameworks where the model is guided to learn representations useful for potential downstream tasks from a set of pretext tasks.
DIM \cite{hjelm2018learning} maximizes the mutual information between input and output. AMDIM \cite{bachman2019learning} makes multiple views of an input and tries to maximize the mutual information between input and output using different views. The key difference between AMDIM and our SDSSL is that while AMDIM encourages the final representation to mimic intermediate features from a particular layer using convolutional neural networks, SDSSL enforces every intermediate representation to mimic the final feature, which is best suited for the pretext task, \textit{i.e., instance discrimination.}
CPC \cite{oord2018representation} trains the representation in a sequential model by using a contrastive method and shows that InfoNCE loss maximizes the lower bound of mutual information between the inputs and the representations. 
SimCLR \cite{chen2020simple} and MoCo \cite{he2020momentum} have shown impressive performance on various benchmarks by combining strong augmentations with contrastive learning objective, but they are yet limited in that they either require huge batch size or a separate memory bank.
BYOL \cite{grill2020bootstrap} has successfully overcome this limitation by only using positive samples while boosting the performance at the same time.
Meanwhile, as transformers~\cite{vaswani2017attention,dosovitskiy2020image} gain increasing popularity in the vision domain, self-supervised learning using transformer has been studied as well~\cite{chen2021empirical,caron2021emerging, ge2021revitalizing}, pushing the previous state-of-the-art further and introducing several beneficial properties absent in traditional CNN models.
Despite steady improvements in quantitative evaluations, little has been said about \textit{how and why} these methods work.
ReLIC \cite{mitrovic2020representation} introduces the causal mechanism to explain SSL and \cite{wang2020understanding} introduces alignment and uniformity for quantitative analysis. We provide alignment-uniformity analysis in Secs. \ref{subsec:analysis} to shed light on the underlying factors that motivate the success of SDSSL.


\noindent \textbf{Knowledge distillation} Knowledge distillation (KD) is one of the regularization methods widely used to improve model performance \cite{hinton2015distilling,NEURIPS2018_6d9cb7de,he2019knowledge}. The conventional offline KD framework utilizes a pre-trained teacher network to provide additional learning signals to the student network that is primarily trained with labels.
In contrast, online KD methods adopt concurrent training scheme, where the teacher and the student are trained simultaneously, distilling information from each other.
Recently, several works have explored the concept of self-distillation, where the knowledge from previous snapshot of the model is distilled to the current one~\cite{radosavovic2018data,rebuffi2017icarl}. Multi-exit~\cite{phuong2019distillation}, among others, enforces the prediction from lower layers to match that of higher layers, which has similarities with our SDSSL in the high level idea.
However, unlike the aforementioned methods, SDSSL is fully self-supervised, not requiring any labels in the training process. We also note that our method differs from others in the formulation of distillation, as ours is driven by instance-discrimination-based SSL objective.

\section{Method}
\subsection{Baselines}
 \noindent \textbf{SimCLR} makes two views, $X_1$ and $X_2$, of an input image $X$ (positive sample), by performing separate random augmentations. 
 The representations of $X_1$ and $X_2$ from the backbone network are first projected, and the contrastive objective enforces the cosine similarity between positive samples to be maximized while minimizing it between negative samples (other images in the batch)~\cite{bridle1990training}.
 
 
 \vspace{1mm}
 
 \noindent \textbf{MoCo v3} learns from the contrastive loss like SimCLR, but instead of using an identical network to generate features for $X_1$ and $X_2$, a teacher network with exponential moving average (EMA) parameters is used. Randomly augmented views $X_1$ and $X_2$ are forwarded to the student network and the teacher network respectively, and then projected. The projected output of the student network is further processed through an additional MLP head to perform contrastive learning.
 
 
 \vspace{1mm}
 
 \noindent \textbf{BYOL} also has an EMA teacher and a predictor like MoCo v3, but learns simply by increasing the cosine similarity of positive samples without using the contrastive loss. Therefore, unlike the aforementioned SSL frameworks that utilize negative samples, the performance is robust to the choice of batch size.
 
 \vspace{1mm}
 
 \noindent \textbf{SSL objective} functions $\mathcal{L}_{ssl}$ differ for different baselines.
 Following common practice, let $q$ denote the output of the student's last MLP head (projector or predictor) and $z$ denote the output of the teacher's (student in SimCLR) projector. Then the objective for BYOL is
\begin{equation}
    \mathcal{L}_{ssl}(q,z)  = 2 - 2 \cdot {\dfrac{\inner{q,z}}{\norm{q}_2 \cdot \norm{z}_2} }
\end{equation}
while for the two contrasitive methods,  

\begin{equation}
    \label{eqn:contrastive}
    \mathcal{L}_{ssl}(q,z)  = {-\log\dfrac{e^{\inner{q, z^+} / \tau}}{e^{ \inner{q, z^+} / \tau} + \sum_{z^-}e^{\inner{q, z^-} / \tau}}}
\end{equation}

where $\inner{\cdot, \cdot}$ denotes the inner product, $\tau$ is a temperature parameter and $z^+/z^-$ are for the positive/negative samples. We note that the $q$ and $z$ are L2-normalized in \cref{eqn:contrastive}

\begin{figure}[t]
    \caption{\footnotesize\textbf{Illustration of SDSSL} in MoCo v3 and BYOL. For SimCLR, predictors don't exist and teacher network is identical to student network. Solid line is updated by $\mathcal{L}_{ssl}$ while dotted line is for $\mathcal{L}_{isd}$ and $\mathcal{L}_{pred}$.\vspace{-2mm}}\label{fig:sdssl}
    {\includegraphics[width=0.95\linewidth]{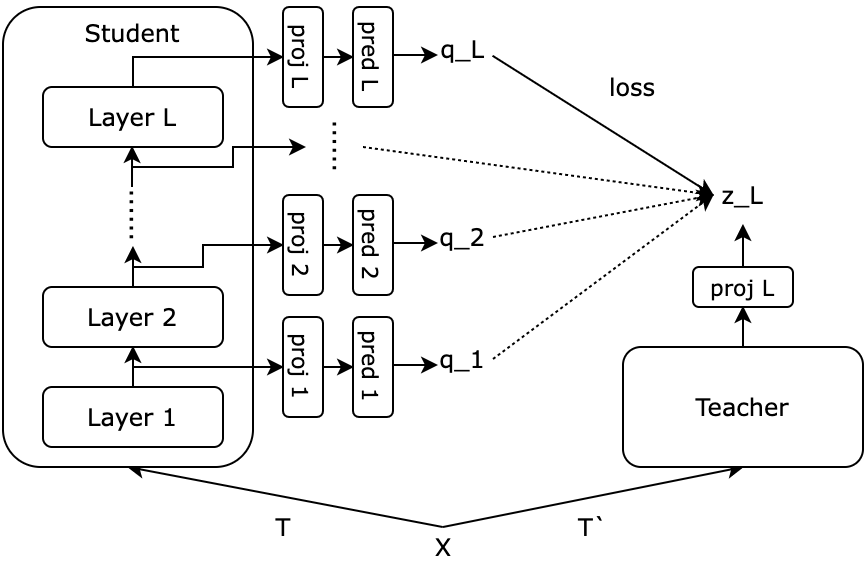}}
\end{figure}

 \subsection{SDSSL}
 \label{method:sdssl}
 We propose Self-Distilled Self-Supervised Representation Learning (SDSSL) which provides explicit signal to the intermediate representation by inducing intermediate representation to mimic the output representation as illustrated in Fig. \ref{fig:sdssl}. Our method can be applied to any existing SSL frameworks that aligns
representations from multi-views. 

\vspace{2mm}

\vspace{2mm}

\noindent
\textbf{Self-Distilled SSL}
We define our \textit{intermediate self-distillation} loss $\mathcal{L}_{isd}$, which tries to maximize the mutual information of the output of an intermediate layer $l$ and the final layer $L$  ($I(f_l^1;f_L^2)$), as the following

\begin{equation}
    \mathcal{L}_{isd} = {1 \over L-1}\sum_{l=1}^{L-1} \mathcal{L}_{ssl}(q_l,\text{sg}(z_L)),
    \label{eq:isd}
\end{equation}
where $q_{l}$ is the representation of the $l^{th}$ layer of the student encoder passed through the MLP heads corresponding to each layer, and $z_L$ is the output of the teacher MLP head.
The stop-gradient operator, $\text{sg}(z_L)$ implies that the gradient is not propagated through $z_L$ so that only $q_l$ learns to predict $z_L$ without affecting $z_L$. 
The objective of SDSSL consists of $\mathcal{L}_{ssl}$ and $\mathcal{L}_{isd}$, resulting in $\mathcal{L}_{sdssl}$:
\begin{equation}
    \mathcal{L}_{sdssl} = \mathcal{L}_{ssl}(q_L,z_L) + \alpha\mathcal{L}_{isd},
    \label{eq:SDSSL}
\end{equation}
where the choice of $\alpha$, which controls the weight of the self-distillation loss, is detailed in Secs. \ref{subsec:implementation}. 

We observe that for frameworks where predictors exist, simply using \cref{eq:SDSSL} leads to some performance improvement, but can be further enhanced.
This is because the predictors of the intermediate layers are only updated using gradients from $\mathcal{L}_{isd}$ as opposed to the encoder, which is able to utilize both $\mathcal{L}_{{isd}}$ and $\mathcal{L}_{{ssl}}$. Consequently, the optimality of the predictors at intermediate layers are not guaranteed, which is a key component of SSL training as discussed by \cite{grill2020bootstrap}. Simply enlarging $\alpha$ causes the last layer of the encoder to be sub-optimal, because this updates the intermediate backbone layers as well. 
To alleviate this issue, we employ another loss $\mathcal{L}_{pred}$: 
\begin{equation}
    \mathcal{L}_{pred} = \sum_{l=1}^{L} \mathcal{L}_{ssl}(\text{pred}(\text{sg}(h_l)),\text{sg}(z_L)),
\end{equation}
where $h_l$ is the representation of the $l^{th}$ layer of the student after passing through the projector. To only update the predictors, the $\text{sg}(\cdot)$ operator is used to $h_l$. By doing so, we attain better predictors for distillation and hence the final objective for SSL frameworks with predictors is  
\begin{equation}
    \Tilde{\mathcal{L}}_{sdssl} = \mathcal{L}_{ssl}(q_L,z_L) + \alpha\mathcal{L}_{isd} + \beta\mathcal{L}_{pred}.
    \label{eq:sdssl_p}
\end{equation}
We use $\beta=1$ in \cref{eq:sdssl_p}. Alg. \ref{alg:code} provides the pseudocode for SD-MoCo v3 which applies SDSSL to MoCo v3.

\begin{algorithm}[t]
\caption{SD-MoCo v3: PyTorch-like Pseudocode}
\label{alg:code}
\definecolor{codeblue}{rgb}{0.25,0.5,0.5}
\definecolor{codekw}{rgb}{0.85, 0.18, 0.50}
\begin{lstlisting}[language=python]
# f_s: student: ViT + projectors
# f_t: momentum teacher: ViT + projector
# p: predictors
# alpha: intermediate self-distillation ratio
# tau: temperature
# L: number of layers in ViT

for x in loader:  # load a minibatch x with N samples
    x1, x2 = aug(x), aug(x)  # random augmentation
    q1, q2 = f_s(x1), f_s(x2) # shape: [L*N, dim]
    z1, z2 = f_t(x1), f_t(x2) # shape: [N, dim]

    loss_pred = ctr(p(q1.detach()), z2, L)
    loss_pred += ctr(p(q2.detach()), z1, L)
    
    q1, q2 = p(q1), p(q2)
    
    q1_isd, q1 = split(q1, [(L-1)*N, N])
    q2_isd, q2 = split(q2, [(L-1)*N, N])
    
    loss_isd = ctr(q1_isd, z2, L-1)
    loss_isd += ctr(q2_isd, z1, L-1)
    
    loss = ctr(q1, z2) + ctr(q2, z1)
    loss += alpha * loss_isd + L * loss_pred
    
    loss.backward()
    optimizer.update(f_s, p)
    momentum_update()
  
# contrastive loss
def ctr(q, z, num_layers=1):
    logits = mm(q, z.t())  # [num_layers*N, N] pairs
    labels = repeat(arange(N), num_layers)
    loss = CrossEntropyLoss(logits/tau, labels) 
    return 2 * tau * loss
\end{lstlisting}
\end{algorithm}

\section{Experiments}

In this section we describe details of our implementation. We follow the convention of MoCo v3 \cite{chen2021empirical} unless otherwise noted. We show that SDSSL outperforms the baselines in various downstream tasks including ImageNet. 
Moreover, we employ t-SNE \cite{van2008visualizing}, Centered Kernel Alignment (CKA)~\cite{cortes2012algorithms,kornblith2019similarity} and uniformity-alignment framework~\cite{wang2020understanding} to analyze the representations of SDSSL in an attempt to demystify its success. As we later show, the key is learning better representations in the lower layers, pushing the overall performance curve upwards. Thorough ablations show the effectiveness of individual components of SDSSL and the comparison of SDSSL with ResNet backbones reconfirms the effectiveness of our method.

\subsection{Implementation Details}
\label{subsec:implementation}

\noindent \textbf{ViT Architecture} We adopt the sine-cosine variant \cite{vaswani2017attention} in 2-D for positional embedding and freeze the random initialized patch projector. We concatenate patch embedding with a learnable [CLS] token and add its positional embedding. The representations are outputs of [CLS] token after passing through each transformer block and the layer normalization layer \cite{ba2016layer}.
\begin{table}[t]
\centering
\begin{adjustbox}{width = 0.98\textwidth}
\begin{tabular}{l|cc|cc|cc}
\hline
\multirow{2}{*}{\textbf{Framework}} & \multicolumn{2}{c|}{\textbf{ViT-S/32}} &
\multicolumn{2}{c|}{\textbf{ViT-S/16}} &\multicolumn{2}{c}{\textbf{ViT-B/16}} \\
                  & k-NN & Linear & k-NN & Linear & k-NN & Linear \\ \hline \hline
                SimCLR & 51.5 & 52.8 & 57.8 & 62.1 & 62.1 & 70.5 \\
                SD-SimCLR & \bf{53.4} & \bf{55.3} & \bf{59.1} & \bf{65.0} & \bf{64.4} &
                \bf{72.1} \\ \hline
                BYOL & 56.4 & 59.8 & 66.0 & 70.3 & 68.1 & 73.7 \\
                SD-BYOL & \bf{57.9} & \bf{61.8} & \bf{67.2} & \bf{71.5} & \bf{70.3} & \bf{{74.5}} \\ \hline
                MoCo v3 & 57.1 & 60.7 & 66.5 & 70.0 & 69.7 & 75.1 \\
                SD-MoCo v3 & \bf{59.0} & \bf{63.7} & \bf{68.0} & \bf{71.5} & \bf{72.0} & \bf{76.0} \\ \hline
\end{tabular}
\end{adjustbox}
\caption{\footnotesize \textbf{ImageNet Evaluation.} Comparison with three competitive baselines and SDSSL. ViT-S/32, ViT-S/16 and ViT-B/16 are trained on ImageNet for 300 epochs. For each framework, ViT-S/32, ViT-S/16 and ViT-B/16 share the same set of hyper-parameters except batch size.
\vspace{-3mm}}
\label{table:main}
\end{table}
\vspace{1mm}

\noindent \textbf{MLP Heads} Following \cite{chen2020simple,grill2020bootstrap}, projectors and predictors are set as 3-layer MLPs and 2-layer MLPs, respectively. Batch normalization \cite{ioffe2015batch} is applied to all output layers except BYOL and the hidden layers for all methods. The dimension of the hidden layer is 4096 for the last projector and all predictors, but 2048 for intermediate projectors. All outputs have 256 dimension. 
For frameworks using an exponential moving average (EMA) teacher, the teacher's projector is updated using the student's projector via EMA. This is done in SDSSL as well using only the last projector. In ablation study, we show that the effects of intermediate projector according to the number of layers.

\vspace{1mm}

\noindent \textbf{Hyper-parameter} We use AdamW \cite{loshchilov2017decoupled} as the optimizer and batch size of 4096 for ViT-S/32 and 1024 for ViT-S/16 \cite{touvron2021training} and ViT-B/16. Learning rate is 1.5e-4 for MoCo v3 and BYOL, 1.3e-4 for SimCLR. We also adopt learning rate warmup for 40 epochs and cosine decay after warmup \cite{goyal2017accurate}. Weight decay is 0.1. For $\alpha$, cosine scheduling \cite{loshchilov2016sgdr} is performed from 0$\sim$1.0.

\subsection{Main Results}
\label{subsec:main}
\noindent \textbf{ImageNet Pretraining} We experiment with ViT-S/32, ViT-S/16 and ViT-B/16 on three self-supervised learning frameworks. In Table \ref{table:main} we validate the representations found in ImageNet \cite{deng2009imagenet} pretrained encoder through using k-NN \cite{wu2018unsupervised} and linear evaluation. We follow the protocol of MoCo v3 for the linear evaluation and DINO for k-NN. Across all frameworks, models, and evaluations, applying SDSSL increases performance.
The baseline accuracies are lower than those reported in MoCo v3 paper \cite{chen2021empirical}, because of using 1024 batch size instead of 4096 due to computation constraint. Contrastive frameworks are particularly affected by the batch size. Nevertheless, our method significantly improves upon our reproduced baselines. 
In ViT-S models, linear evaluation performance improved more than or equal to it did on k-NN, whereas in ViT-B/16 model, k-NN performances improved more.
We used 8 and 4 NVIDIA A100 for five days to train our ViT-B/16 and ViT-S/16 models respectively, and 4 NVIDIA A6000 for three days to train ViT-S/32 models.

\vspace{1mm}

\begin{figure}[t]
    {\caption{\footnotesize \textbf{Multi-exit.} Linear evaluations on Imagenet for baselines and SDSSL on each layer using pretrained ViT-S/16 on ImageNet for 300 epochs. SDSSL methods outperform the corresponding baselines at all layers and shows less degradation for earlier layers.}\label{fig:multi}}
    {\includegraphics[width=0.98\linewidth]{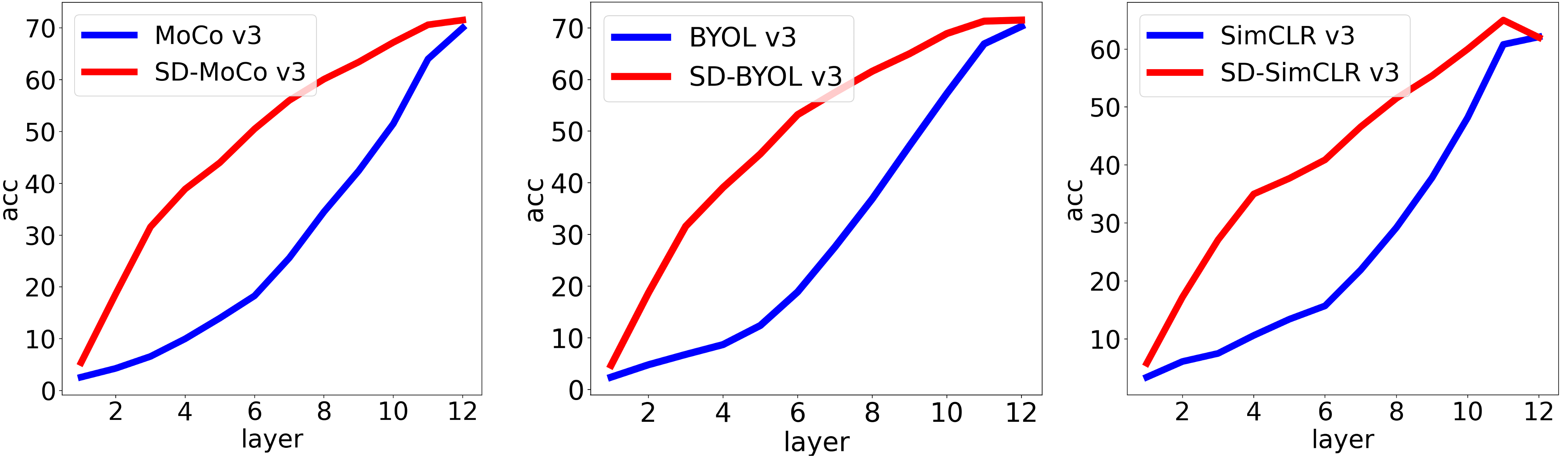}}
\end{figure} 

\noindent \textbf{Multi-exit}
Since self-distillation enables lower layers to learn from the higher layers, we expect that the lower layers of SDSSL learned more meaningful representations than those of the baselines. This is verified in Figure \ref{fig:multi}, which shows that the representations of lower layers for SDSSL are much more suitable as features than the counterparts. We performed linear evaluation on ImageNet using the frozen representations of each layer. In the last layer, the accuracy increased by 1.2\%p, and the 6th layer showed the largest performance gap of 34.3\%p in MoCo v3. Similar phenomena occur in BYOL and SimCLR.

\subsection{Transferability}
\label{sec:transfer}
In this subsection, we evaluate the transferability of our method on various downstream tasks. 
Following DINO \cite{caron2021emerging}, we evaluate on the image retrieval task. In addition, we also evaluate on the copy-detection task and the video segmentation task, which uses features of patches rather than the [CLS] token. The three evaluation protocol do not require additional training of the encoder.
Then, we evaluate on other image classification datasets such as CIFAR-10, CIFAR-100 \cite{krizhevsky2009learning}, Oxford Flowers-102 \cite{nilsback2008automated}, Oxford-IIT-Pets \cite{parkhi2012cats}, CUB~\cite{WelinderEtal2010}, AirCraft~\cite{maji13fine-grained}, Cars~\cite{KrauseStarkDengFei-Fei_3DRR2013}, Dogs~\cite{KhoslaYaoJayadevaprakashFeiFei_FGVC2011}, NABirds~\cite{7298658} and ImageNet by k-NN, linear evaluation and end-to-end fine-tuning \cite{dosovitskiy2020image}. Experiments are performed using all the three frameworks with ViT-S/16.

\vspace{1mm}


\noindent \textbf{Image Retrieval} 
Revisited \cite{radenovic2018revisiting} Oxford and Paris image retrieval datasets \cite{philbin2008lost} contain 3 splits of various difficulty with query and database pairs. We evaluate all baselines and SDSSL on the Medium and Hard splits. We directly apply k-NN for image retrieval. As shown in Table \ref{table:ImgR}, SDSSL outperform baselines.

\vspace{1mm}

\begin{table}
\resizebox{0.78\linewidth}{!}{\scriptsize
\begin{tabular}{lcccc}
\hline
\multirow{2}{*}{Framework} & \multicolumn{2}{c}{ROx} & \multicolumn{2}{c}{RPar} \\
                  & M & H & M & H \\ \hline \hline
                SimCLR & 20.1 & 3.9 & \bf{42.8} & \bf{15.3} \\
                SD-SimCLR & \bf{22.0} & \bf{5.0} & 42.3 & 15.1 \\ \hline
                BYOL & 27.7 & 6.9 & 51.7 & 22.2 \\
                SD-BYOL & \bf{28.5} & \bf{7.7} & \bf{52.0} & \bf{22.5} \\ \hline
                MoCo v3 & 26.3 & 6.4 & 51.0 & 21.9 \\
                SD-MoCo v3 & \bf{26.7} & 6.4 & \bf{52.4} & \bf{22.7} \\ \hline
\end{tabular}
}
{\caption{\footnotesize \textbf{Image Retrieval.} Comparison of performance between baseline and SDSSL on image retrieval task. ViT-S/16 are pre-trained using each framework on ImageNet for 300 epochs. We evaluate image retrieval task using k-NN.}
\label{table:ImgR}
}
\end{table}

\begin{table}
\resizebox{0.9\linewidth}{!}{\scriptsize
\begin{tabular}{lcccc}
\hline
\multirow{2}{*}{Framework} & \multicolumn{1}{c}{Copy D.} & \multicolumn{3}{c}{Video S.} \\
& mAP & $(\mathcal{J}\&\mathcal{F})_m$ & $\mathcal{J}_m$ & $\mathcal{F}_m$ \\ \hline \hline
SimCLR & 74.7 & 61.8 & 59.9 & 63.6 \\
SD-SimCLR & \bf{75.5} & \bf{62.1} & \bf{60.3} & \bf{64.0} \\ \hline
BYOL & \bf{74.5} & 60.2 & 58.1 & 62.3 \\
SD-BYOL & 74.2 & \bf{60.9} & \bf{59.0} & \bf{62.7} \\ \hline
MoCo v3 & 74.8 & 62.0 & 60.2 & 63.8 \\
SD-MoCo v3 & \bf{76.3} & \bf{62.1} & \bf{60.4} & \bf{63.9} \\ \hline
\end{tabular}}
{\caption{\footnotesize\textbf{Copy detection and video segmentation.} For all scores, higher means better. The reported scores are the performance of the best layer in each method. ImageNet pretrained ViT-S/16 models are used to evaluate.\vspace{-4mm}}
\label{tab:detect}
\vspace{-1mm}
}
\end{table}

\begin{figure}[t]
    \caption{\footnotesize\textbf{Copy Detection and Video Segmentation.} (Left) Results of copy detection and video segmentation tasks on MoCo v3 and SD-MoCo v3 for each layer. With the exception of the some layers, SD-MoCo v3 outperforms MoCo v3. The best performing layers are 7th for SD-MoCo v3 and 10th for MoCo v3 in copy detection and video segmentation. The trends are almost the same for BYOL/SD-BYOL (middle) and SimCLR/SD-SimCLR (right).\vspace{-4mm}}
    {\includegraphics[width=1.0\linewidth]{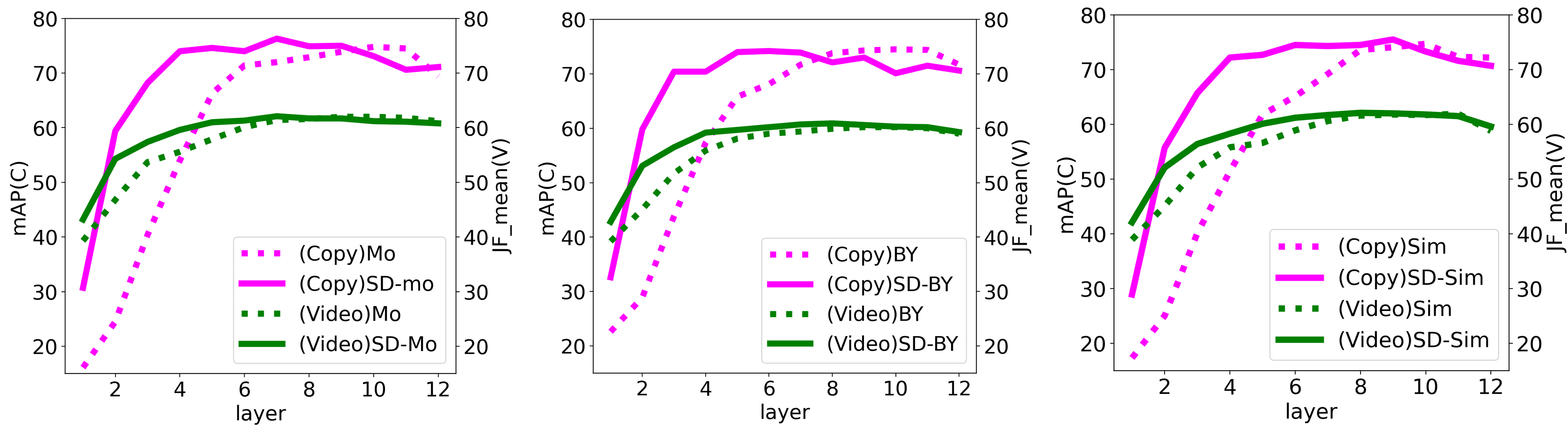}}
    \vspace{-5mm}
    \label{fig:copy}
\end{figure}

\begin{table*}[t]
\begin{adjustbox}{width = 0.85\linewidth}
{\scriptsize
\begin{tabular}{l|cccccccccc|c}
\hline
Framework & CI-10 & CI-100 & Flower & Pets & CUB & ACraft & Cars & Dogs & NABirds & INet & Avg. \\ \hline \hline
SimCLR-knn & \textbf{87.2} & 65.7 & 77.8 & 71.8 & 28.5 & \textbf{22.9} & 15.6 & 50.1 & 17.3 & 57.8 & 49.5\\
 & 86.9 & \textbf{66.0} & \textbf{79.1} & \textbf{73.9} & \textbf{29.2} & 22.8 & \textbf{15.7} & \textbf{52.8} & \textbf{18.2} & \textbf{59.1} & \bf{50.4}\\ \hline
linear & 81.9 & 58.4 & 80.4 & 71.4 & 39.8 & \textbf{24.3} & \textbf{17.5} & 58.9 & 31.4 & 62.1 & 52.6\\
 & \textbf{82.8} & \textbf{59.0} & \textbf{81.6} & \textbf{74.2} & \textbf{41.9} & 22.6 & 16.9 & \textbf{62.2} & \textbf{32.5} & \textbf{65.0} & \bf{53.9}\\ \hline
fine & 97.8 & 86.1 & 95.8 & 88.0 & 73.6 & 75.9 & 85.1 & 72.5 & 68.7 &  78.6 & 82.2\\
 & \bf{98.5} & \textbf{88.6} & \textbf{96.8} & \textbf{89.9} & \textbf{74.2} & \bf{81.2} & \bf{87.0} & \bf{80.0} & \textbf{69.3} & \bf{78.7} & \bf{84.4}\\ \Xhline{2\arrayrulewidth}
BYOL-knn & 90.0 & 70.6 & \textbf{85.2} & 83.4 & 52.8 & 31.1 & 19.9 & 67.0 & 38.2 & 66.5 & 60.4\\
 & \textbf{91.5} & \textbf{72.5} & 85.0 & \textbf{85.3} & \textbf{54.5} & \textbf{32.4} & \textbf{21.5} & \textbf{69.0} & \textbf{40.7} & \textbf{68.0} & \bf{62.0} \\ \hline
linear & 89.9 & 73.4 & \textbf{92.7} & 87.5 & 70.7 & 46.4 & 42.3 & 76.8 & 61.6 & 70.3 & 71.2\\
 & \textbf{92.4} & \textbf{75.2} & 92.6 & \textbf{88.0} & \textbf{73.4} & \textbf{46.6} & \textbf{43.8} & \textbf{78.5} & \textbf{63.2} & \textbf{71.5} & \bf{72.5}\\ \hline
fine & 98.6 & 89.3 & \textbf{97.4} & 91.1 & 78.9 & 80.4 & 88.7 & 80.9 & 74.9 & 79.3 & 86.0\\
 & \textbf{98.9} & \textbf{89.4} & 97.2 & \textbf{91.4} & \textbf{79.9} & 80.4 & \textbf{88.9} & \textbf{81.1} & \textbf{75.7} & \textbf{79.5} & \bf{86.2}\\ \Xhline{2\arrayrulewidth}
MoCo v3-knn & \textbf{91.8} & \textbf{73.8} & 85.4 & 83.6 & 51.2 & 30.4 & 21.3 & 67.7 & 36.0 & 66.0 & 60.7\\
 & 91.2 & 73.4 & \textbf{85.5} & \textbf{84.4} & \textbf{53.1} & \textbf{32.5} & \textbf{22.6} & \textbf{69.4} & \textbf{38.5} & \textbf{67.2} & \bf{61.8}\\ \hline
linear & 90.1 & 73.9 & 92.6 & \textbf{87.6} & 70.6 & \textbf{47.3} & 41.2 & 77.7 & 59.6 & 70.0 & 71.1\\
 & \bf{90.2} & \textbf{74.4} & 92.6 & 87.5 & \textbf{71.7} & 47.2 & \textbf{43.8} & \textbf{78.3} & \textbf{61.6} & \textbf{71.5} & \bf{71.9}\\ \hline
fine & 98.7 & \textbf{89.5} & 97.2 & 90.9 & 78.5 & \textbf{81.6} & 86.8 & 78.8 & 74.1 & 79.4 & 85.6\\
 & 98.7 & 89.2 & \textbf{97.3} & \textbf{91.4} & \textbf{79.3} & 81.3 & \textbf{87.8} & \textbf{79.7} & \textbf{75.8} & \textbf{79.6} & \bf{86.0}\\ \hline
\end{tabular}}
\end{adjustbox}
\caption{\footnotesize \textbf{Classification.} We report k-NN, linear and fine-tuning performances for 10 classification datasets. The upper row is the baseline accuracy, and the lower row is the SDSSL accuracy.\vspace{-3mm}}
\label{table:Cls}
\end{table*}

\noindent \textbf{Copy-detection} 
We report the mean average precision (mAP) of copy-detection on the \textit{strong} subset of INRIA Copydays dataset \cite{douze2009evaluation}.
The goal of Copy-detection is to recognize the original image when given a distorted (e.g. blur, insertion, print, scan) version of it.  
Following \cite{berman2019multigrain}, we use the 10K samples of the YFCC100M dataset \cite{thomee2016yfcc100m} as distractors, while 20K samples are used for whitening \cite{berman2019multigrain} the features. The features of [CLS] token and patch token are pooled using GeM \cite{radenovic2018fine} and concatenated. 
We use features of all layers to verify whether similar trend occurs as in the multi-exit experiment. We observe in Figure \ref{fig:copy} that most SD-MoCo v3 intermediate features surpass those of MoCo v3 and have better performance on the respective best-performing features. For SD-MoCo v3 and MoCo v3, this is the 7th and 10th layer, respectively. We believe that the best-performing features are not formed in the final layer for some tasks that utilizes the features of the patch rather than using only the [CLS] token. 
Moreover, for SDSSL the best-performing layer is formed in the lower layers than the baseline. For SD-SimCLR and SimCLR, the best-performing layers are 9th and 10th, respectively; for
SD-BYOL and BYOL, 6th and 10th layer, respectively.
This may be explained by our knowledge distilling method which intends to extract more information in the lower layers. By providing an explicit loss, our method forms a suitable 
feature for copy detection earlier in the lower layers than the baseline.

\vspace{1mm}

\noindent \textbf{Video segmentation} We perform video instance segmentation on the DAVIS-2017 dataset \cite{pont20172017}. We follow the experimental protocol in Jabri \textit{et al}. \cite{jabri2020space} and segment scenes with a nearest neighbor between consecutive frames in DINO. When all representations of all layers are tested as done in copy detection, a similar trend is observed in the video segmentation task as well. The best performing layer is 7th and 10th for SD-MoCo v3 and MoCo v3, respectively, and SD-MoCo v3 outperforms MoCo v3 as shown in Table \ref{tab:detect}. For SD-SimCLR and SimCLR, the best-performing layers are 8th and 9th, respectively; for SD-BYOL and BYOL, 8th and 10th layer, respectively.


\noindent \textbf{Classification} 
In this section, we demonstrate the results for image classification on CIFAR-10, CIFAR-100, Oxford Flowers-102, Oxford-IIT-Pets, CUB, AirCraft, Cars, Dogs, NABirds and ImageNet. 
Since end-to-end fine-tuning could lead to over-fitting on the specific dataset, this may obscure the evaluation of representation quality of the pre-trained encoder \cite{radford2021learning}.
To address this issue,
we also report scores from k-NN and linear evaluation. 
In Tab. \ref{table:Cls}, we witness improvements from SDSSL on most of the datasets for all three baselines. Performance gains in k-NN and linear evaluation are especially noticeable, which implies that representations from SDSSL are more separable in the feature space. Further analysis on the representations will be presented in Secs. \ref{subsec:analysis}. Results from end-to-end fine-tuning displays the typical \textit{diminishing gains} as the overall baseline performance has been greatly boosted.


\begin{figure*}[t]
    \centering
    \includegraphics[width=0.9\linewidth]{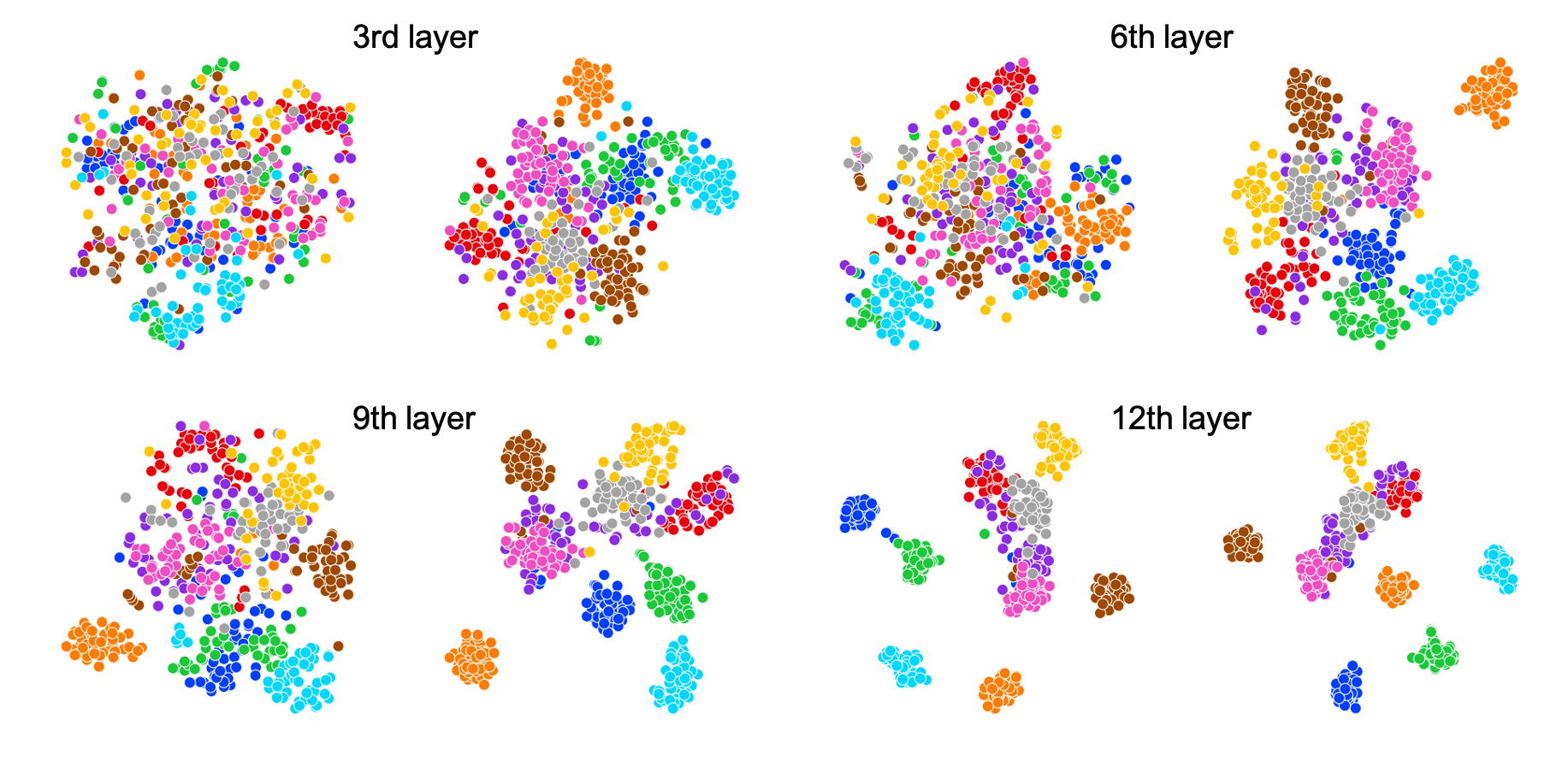}
    \caption{\footnotesize \textbf{Representation Visualization.} The representations of each layer of MoCo v3 and SD-MoCo v3 are visualized using t-SNE. The left side of each layer is MoCo v3, and the right side is SD-MoCo v3. 10 random classes of ImageNet validation set are drawn. We observe that the representations of the lower layers are aggregated better by class when SDSSL is applied.\vspace{-4mm}}
    \vspace{-4mm}
    \label{fig:tsne}
\end{figure*}

\begin{figure}[t]
    \centering
    \includegraphics[width=0.98\linewidth]{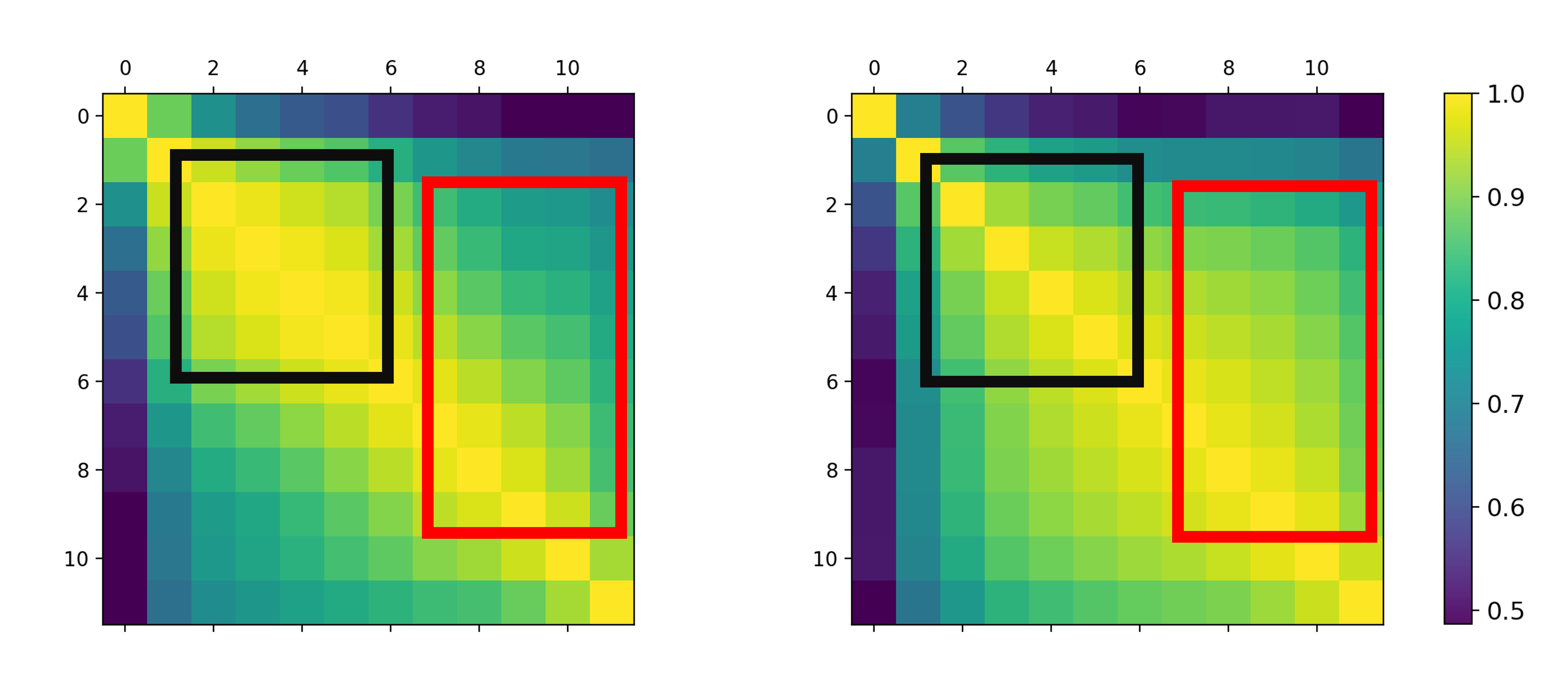}
    \caption{\footnotesize\textbf{Representation Similarity.} We compute CKA heatmap for all pairs of layers for MoCo v3 \textit{(left)} and SD-MoCo v3 \textit{(right)}.
    Black box demonstrates SDSSL's reduced similarity between neighboring layers and red box shows that lower layers faithfully mirrors upper layers.
    \vspace{-3mm}
    }
    \vspace{-7mm}
    \label{fig:cka}
\end{figure}

\subsection{Analysis}
\label{subsec:analysis}
\noindent\textbf{Qualitative analysis} Fig. \ref{fig:tsne} displays t-SNE visualization of representations from 10 randomly selected classes of ImageNet validation set, where SDSSL on MoCo v3 \textit{(right)} shows clearly more separable representations compared to the MoCo v3 baseline \textit{(left)}. As one of the basic assumptions of contrastive learning framework is to obtain representations useful for downstream tasks through instance discrimination, this implies that self-distillation encourages the model to form better representations from lower layers, making the overall self-supervised learning task easier. Results in Fig. \ref{fig:multi} and Fig. \ref{fig:copy} support this claim.

CKA~\cite{cortes2012algorithms,kornblith2019similarity} in Fig. \ref{fig:cka} illustrates the representation similarities between layers. For the MoCo v3 baseline, representations from neighboring layers display high similarity \textit{(black box, left)} while the scores are largely reduced for distant layers \textit{(red box, left)}. SDSSL, on the other hand, shows more uniform similarity structure \textit{(right)} with lower local similarity and higher global similarity. As previous work~\cite{raghu2021vision} points at the uniform representation similarities across layers as a key property that distinguishes vision transformers from convolutional networks, we conjecture this can be one of the drivers for the success of SDSSL reinforcing the characteristics of ViTs.

\noindent\textbf{Quantative analysis}
Wang \textit{et al.} \cite{wang2020understanding} demonstrated that contrastive learning optimizes two distinct metrics: (1) \textit{alignment}, which quantifies compactness of representations of \textit{positive} samples
\begin{equation}
    \mathcal{L}_{ali}(f;\gamma) \triangleq \mathop{\mathbb{E}}_{(x, y)\sim p_{pos}}[\norm{f(x)-f(y)}_2^\gamma], \;\gamma>0,
\end{equation}
for some $\gamma$,
and (2) \textit{uniformity}, which measures how dispersed the entire representations are in a hypersphere using the Gaussian potential kernel \cite{cohn2007universally,borodachov2019discrete}
\begin{equation}
    \mathcal{L}_{uni}(f;t) \triangleq \log\mathop{\mathbb{E}}_{(x, y)\sim p_{data}}[e^{-t\norm{f(x)-f(y)}_2^2}], \;t>0.
\end{equation}

Here, $p_{pos}$ is the distribution of positive pairs generated by random augmentation from the input data, and $p_{data}$ is the overall data distribution.
They asserted that low alignment signifies that the positive samples are close to each other while low uniformity signifies that the negative samples are further apart. Thus, low alignment and low uniformity lead to a better representation with high linearly separability, although the two metrics are inherently in a trade-off relationship. 

Empirically, we observed that SD-MoCo v3 has lower alignment in lower layer, but higher uniformity than vanilla MoCo v3. In the higher layer, the pattern is reversed. In other words, while distances of the positive and negative samples are both close in lower layer, in the higher layer, both are further away. Considering their conflicting characteristics, it is difficult to ascertain which representation is better. 
To answer this question, 
we propose a new metric, uniformity-alignment ratio, that divides the uniformity by alignment.
For formulational simplicity, we compute negative alignment, \textit{i.e.,} the distance between negative samples, as following and divide it by the original alignment, \textit{i.e.,} the distance between positive samples. This successfully removes potential scale ambiguities and indicates the relative distance between negative samples compared to positive samples in an intuitive manner.
\vspace{-1mm}


\begin{figure}[t]
    \centering
    \includegraphics[width=1.0\linewidth]{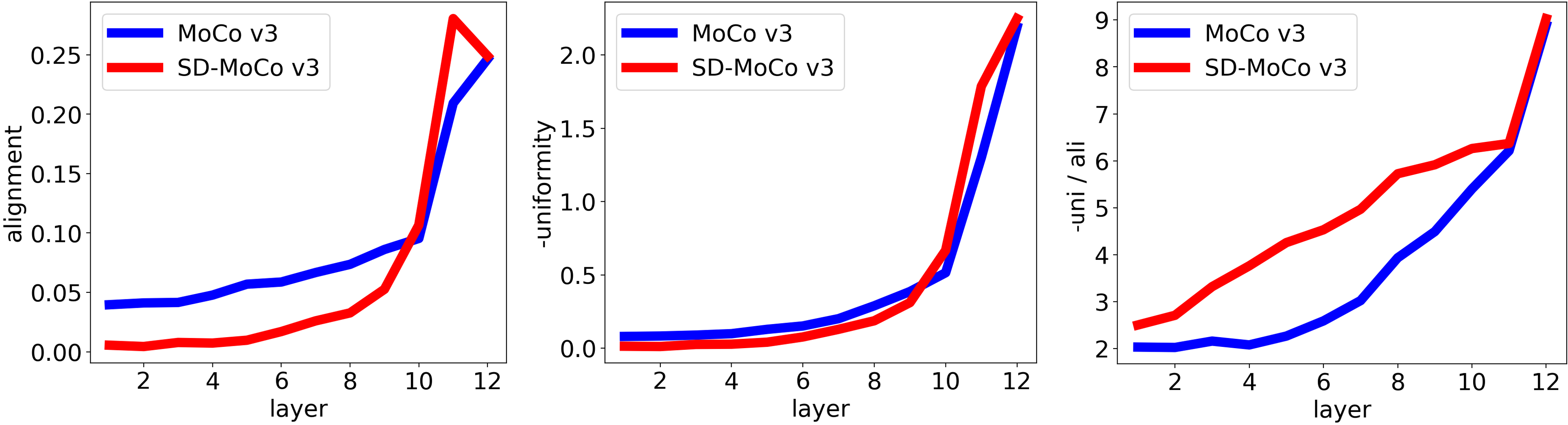}
    \vspace{-1mm}
    {\caption{\footnotesize\textbf{Alignment and Uniformity} measured at each layer of MoCo v3 and SD-MoCo v3 on ImageNet validation set. Because uniformity and alignment have different signs due to the logarithm of uniformity, we report $-\mathcal{L}_{uni}$ for consistency. Additionally, we compute $-\mathcal{L}_{uni}$ / $\mathcal{L}_{ali}$ to estimate efficiency of representation space.
    }\label{fig:ali_uni}}
    \vspace{-4mm}
\end{figure}

\begin{figure}[t]
    \centering
    \includegraphics[width=0.9\linewidth]{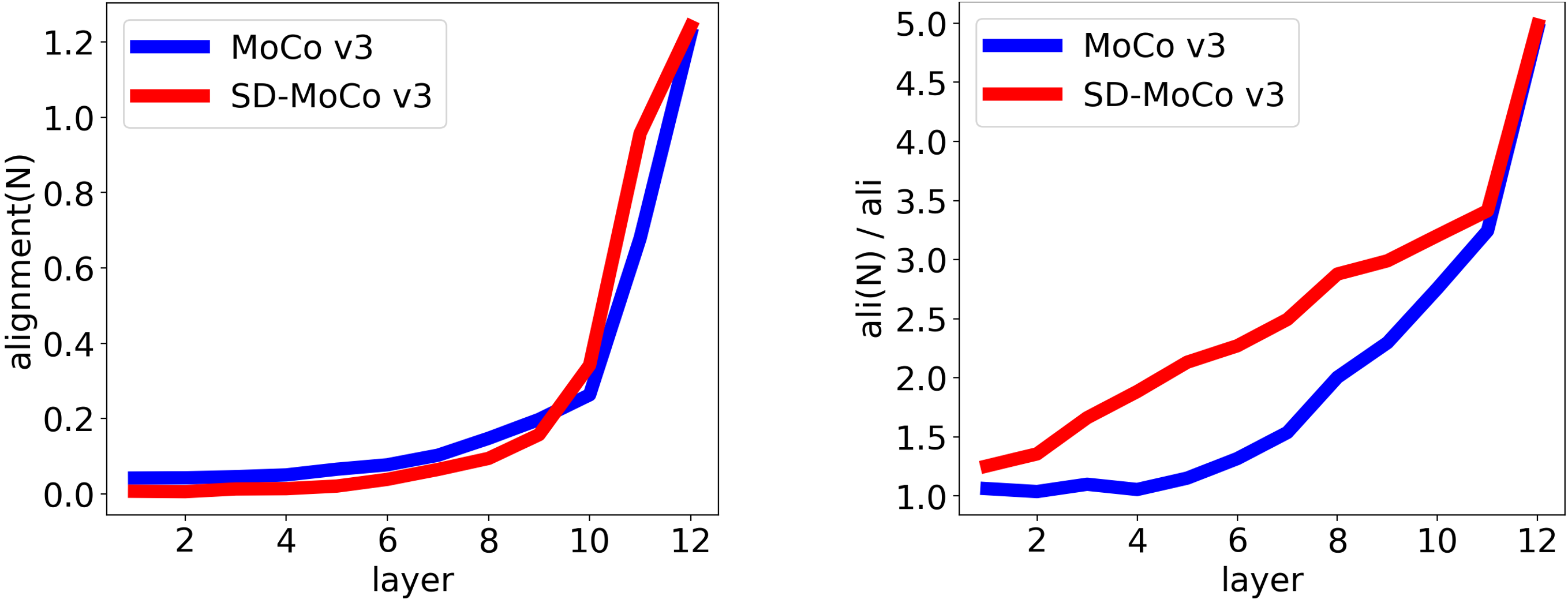}
    \caption{\footnotesize\textbf{Negative Alignment.} We plot negative alignment and alignment ratio $R$ using Imagenet pretrained MoCo v3 and SD-MoCo v3. Negative alignment and alignment ratio show similar pattern with uniformity and uniformity alignment ratio, respectively.
    \vspace{-4mm}
    }
    \vspace{-4mm}
    \label{fig:n_ali_ratio}
\end{figure}

\begin{equation}
    \mathcal{L}_{ali}^n(f;\gamma) \triangleq \mathop{\mathbb{E}}_{(x, y)\sim p_{data}}[\norm{f(x)-f(y)}_2^\gamma], \;\gamma>0.
\end{equation}
Higher $\mathcal{L}_{ali}^n$ means that the negative samples are further apart from each other similar to uniformity.


%
The ratio of negative alignment divided by positive alignment $\mathcal{R} \triangleq \mathcal{L}_{ali}^n / \mathcal{L}_{ali}$ then quantifies how far apart the mean distance between positive and negative samples are. As shown in Figure \ref{fig:n_ali_ratio}, SD-MoCo v3 has higher alignment ratio than MoCo v3 in all layers like uniformity alignment ratio shown in the third column of Fig. \ref{fig:ali_uni}. 
Intuitively, learning a representation space where negative samples are placed further apart from each other compared to positive samples is one of the key desiderata of contrastive learning, and we believe this explains the effectiveness of SDSSL at least partially. Qualitative analyses such as Fig. \ref{fig:tsne} also support this view, delivering a consistent message as to how self-distillation helps self-supervised learning.


\subsection{Ablation Study}
In this subsection, we show the efficacy of ratio scheduling and the predictor loss through ablation and verify these are necessary factors for optimal performance. Furthermore, we experiment how the performance of SDSSL changes according to the values of $\alpha$ and $\beta$.

\begin{figure}[t]
\begin{floatrow}
\begin{adjustbox}{width = 0.5\linewidth, margin=0ex 0ex 0ex 0ex}
\ffigbox{
\includegraphics[width=0.9\linewidth]{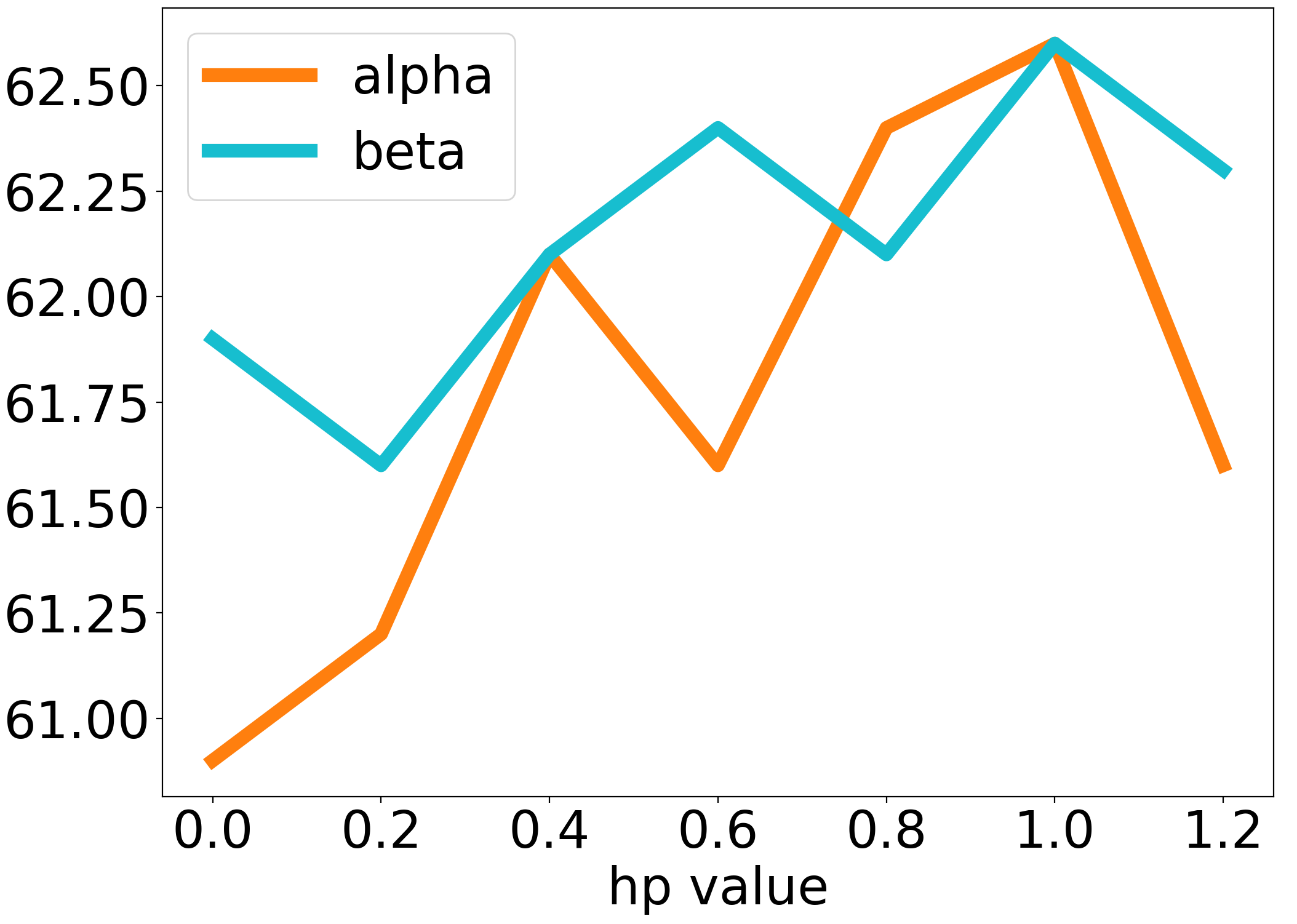}
}
{\captionsetup{width=.9\linewidth, font={Small}}
\caption{\textbf{Hyperparameter sweeping.} We vary $\alpha$ and $\beta$ to see their effect on the ImageNet k-NN accuracy on MoCo v3. When sweeping for a parameter, the other parameter is fixed to 1.0.}\label{fig:alphabeta}}
\end{adjustbox}

\begin{adjustbox}{width = 0.5\linewidth, margin=0ex 0ex 0ex 0ex}
\capbtabbox{
\resizebox{0.85\linewidth}{!}{\scriptsize
\begin{tabular}{lc}
    \hline
        & k-NN \\ \hline \hline
        MoCo v3 & 60.5 \\
        \;\;+pred. loss & 60.8 (+0.3) \\ \hline
        SD-MoCo v3 & 62.6 \\
        \;\;-ratio anneal & 60.2 (-2.4) \\
        \;\;-pred. loss & 61.9 (-0.7) \\ \hline
    \end{tabular}}}
    {\captionsetup{width=0.9\linewidth, font=Small}
    \caption{\textbf{Ablation.} 
    This shows that ratio scheduling is indispensable for SDSSL and pred. loss is also beneficial to the final performance.
    }\label{tab:abl}}
\end{adjustbox}

\end{floatrow}
\vspace{-2mm}
\end{figure}

Tab. \ref{tab:abl} shows that the performance of ablating the predictor loss results in a performance degradation. As discussed, this is consistent with the results in \cite{grill2020bootstrap}, which states that optimality of the predictor is crucial.   
Additionally, when only the predictor loss is used without the intermediate distillation loss $\mathcal{L}_{pred}$, the performance gain with respect to MoCo v3 is minimal (+0.3\%p). This verifies that the intermediate distillation loss is a key component.
During training, we used ratio annealing in \cref{eq:SDSSL} and \cref{eq:sdssl_p}, \textit{i.e.}, $\alpha$ is set very low at initial iterations and gradually increased afterward rather than using a fixed $\alpha$ for the entire training. Without ratio annealing the performance decreases significantly, which shows that self-distilling once some training has been done is important.

Fig. \ref{fig:alphabeta} shows the performance change across ranges of $\alpha$ and $\beta$. For both parameters, the performances generally increase until reaching 1. As discussed, $\alpha$, which controls $L_{isd}$, has a larger impact on the performance than $\beta$. 
In Tab. \ref{tab:abl2} we discuss the effects of some of the design choices of the projector and $\mathcal{L}_{isd}$. Distilling the [CLS] token directly without using a projector (w/o MLP) does not lead to any performance gain, while using a smaller MLP (2-layer) leads to a smaller performance gain. In addition, when $\mathcal{L}_{isd}$ is applied every $n$ layer (MLP/$n$) the performance gain is smaller than applying it every layer as done in SDSSL (MLP/1).

\subsection{SDSSL on ResNet}

We further apply SD-Moco v3 framework on top of ResNet-50 backbone and present evaluations for 200 epochs of training. For our proposed self-distillation, we apply the average pooling to the output activation of each residual block and forward it through an MLP head to compute $\mathcal{L}_{sdssl}$. Visible from \cref{tab:resnet}, we observe significant performance gains on ResNet backbone as well, which is consistent with findings of prior works that incorporate self-distillation in supervised learning~\cite{zhang2019your,phuong2019distillation}. Also, we note that for simplicity, we optimize the self-distillation objective only for the output activations of each residual block, resulting in fewer distillation terms per batch. We expect further performance gains from applying self-distillation to more intermediate activations as in \cref{tab:abl2}, but leave it for future work.

\begin{figure}[t]
\begin{floatrow}
    
\begin{adjustbox}{width = 0.35\linewidth}
\capbtabbox{
\resizebox{0.95\linewidth}{!}{\scriptsize
\begin{tabular}{lc}
    \hline
        & k-NN \\ \hline \hline
        MoCo v3 & 60.5 \\ \hline
        W/o MLP & 60.3 (-2.3) \\
        2L-MLP & 61.3 (-1.3)\\ 
        MLP/4 & 61.3 (-1.3)\\
        MLP/3 & 61.5 (-1.1)\\
        MLP/2 & 61.5 (-1.1)\\
        MLP/1 & 62.6 \\ \hline
    \end{tabular}}}
    {\captionsetup{width=.95\linewidth}
    \caption{\textbf{Ablation on design choices.} 
    We explore the effect of MLP design.
    }\label{tab:abl2}}
\end{adjustbox}

\begin{adjustbox}{width = 0.65\linewidth}
\capbtabbox{
\resizebox{1.0\linewidth}{!}{\scriptsize
\begin{tabular}{lccccc}
\hline
\multicolumn{1}{c}{\multirow{2}{*}{}} & \multirow{2}{*}{k-NN} & \multicolumn{4}{c}{Linear} \\
\multicolumn{1}{c}{}                  &                       & 1st  & 2nd  & 3rd  & last  \\ \hline \hline
MoCo v3 & 60.5 & 10.5 & 23.7 & 45.3 & 60.3 \\
SD-MoCo v3 & \bf{61.2} & \bf{24.7} & \bf{43.5} & \bf{59.2} & \bf{62.2} \\ \hline
\end{tabular}}}
    {\captionsetup{width=.95\linewidth, font={Ssmall}}
    \caption{\textbf{SDSSL on ResNet.} We train ResNet-50 on ImageNet for 200 epochs using MoCo v3.}
    \label{tab:resnet}}
\end{adjustbox}

\end{floatrow}
\vspace{-5mm}
\end{figure}
\section{Discussion}
As SDSSL computes the self-distillation loss through additional MLP heads, there are increase in the memory footprint and computational cost for model training. Upon naive implementations with ViT-B/16, the memory footprint rises by less than 6\% and training time by about 16.9\%. However with parallel computing, the increase in wall clock time can be reduced to 8.6\%, which we believe to be moderate. As this additional cost is independent of the model size, the relative burden becomes smaller as we use larger backbone models with smaller patch size, which is often the case in recent practices.


\section{Conclusion}

In this work, we proposed a self-distillation method generally applicable to existing self-supervised learning frameworks. 
Our method is motivated by the hypothesis that self-distillation between lower representations and output representation may be favorable for separable representation learning, and through experiments, we empirically validated the effectiveness of our method.
We showed SDSSL leads to superior performance not only in the final layers, but also in lower layers through the multi-exit experiment.

\noindent
\textbf{Acknowledgement} This work was supported by NRF grant (2021R1A2C3006659) and IITP grant (No.2022-0-00953), both funded by the Korea government (MSIT).

\clearpage

{\small
\bibliographystyle{ieee_fullname}
\bibliography{egbib}
}

\end{document}